\documentclass{article}
\PassOptionsToPackage{numbers, sort&compress}{natbib}
\usepackage[preprint]{conference_2025}

\usepackage{microtype}
\usepackage{hyperref}
\usepackage{url}
\usepackage{booktabs}
\usepackage{tcolorbox}
\usepackage{lineno}
\usepackage{enumitem}

\usepackage[textsize=tiny]{todonotes}

\newcommand{\CS}{C\&S}

\usepackage{caption}
\usepackage{subcaption}
\usepackage{linguex}
\usepackage{latexsym}
\usepackage[T1]{fontenc}
\usepackage{microtype}
\usepackage{inconsolata}
\usepackage{amsmath}
\usepackage{amssymb}
\usepackage{xspace}
\usepackage{amsthm}
\usepackage{graphicx}
\usepackage{mathtools}
\usepackage{listings}
\usepackage{tikz}
\usepackage{siunitx}
\usepackage{tipa}
\usepackage{float}
\usepackage{bbm}
\usepackage{utils}
\usepackage{tikz-dependency}
\usepackage{booktabs, multirow}

\Crefname{figure}{{Fig.}}{{Figs.}}
\crefname{section}{§}{§§}
\Crefname{section}{§}{§§}
\Crefname{appendix}{{Appendix}}{{Appendices}}

\usepackage{pifont}
\usepackage{tikz}
\usepackage{comment}
\usepackage{dcolumn}
\definecolor{correctColor}{rgb}{0.47, 0.62, 0.85}

\makeatletter

\definecolor{reflectcolor}{HTML}{E97132}
\definecolor{withincolor}{HTML}{CE7DCA}
\definecolor{acrosscolor}{HTML}{808080}

\newcommand{\reflect}{{\color{reflectcolor}{\textbf{self-reflection}}}\xspace}
\newcommand{\within}{{\color{withincolor}{\textbf{within-model prediction}}}\xspace}
\newcommand{\across}{{\color{acrosscolor}{\textbf{across-model prediction}}}\xspace}
\newcommand{\factual}{\texttt{\textcolor{blue}{factual}}\xspace}
\newcommand{\crazy}{\texttt{\textcolor{red}{crazy}}\xspace}

\title{Privileged Self-Access Matters for Introspection in AI}

\author{%
  Siyuan Song\textsuperscript{1} \\
  \texttt{siyuansong@utexas.edu}
  \And
  Harvey Lederman\textsuperscript{1} \\
  \texttt{harvey.lederman@utexas.edu}
  \AND
  Jennifer Hu\textsuperscript{2}\thanks{Co-senior authors.} \\
  \texttt{jennhu@jhu.edu}
  \And \quad \quad \quad
  Kyle Mahowald\textsuperscript{1}\footnotemark[1] \\
  \quad \quad \quad\texttt{kyle@utexas.edu}
  \AND
  \normalfont\small
  \textsuperscript{1}The University of Texas at Austin \quad
  \textsuperscript{2}Johns Hopkins University
}

\begin{document}

\maketitle

\begin{abstract}
Whether AI models can introspect is an increasingly important practical question. But there is no consensus on how introspection is to be defined. Beginning from a recently proposed ``lightweight'' definition, we argue instead for a thicker one.
According to our proposal,  introspection in AI is any process which yields information about internal states through a process more reliable than one with equal or lower computational cost available to a third party.
Using experiments where LLMs reason about their internal temperature parameters, we show they can appear to have lightweight introspection while failing to meaningfully introspect per our proposed definition. 
\end{abstract}

\begin{figure}[htbp]
    \centering
    \includegraphics[width=.7\linewidth]{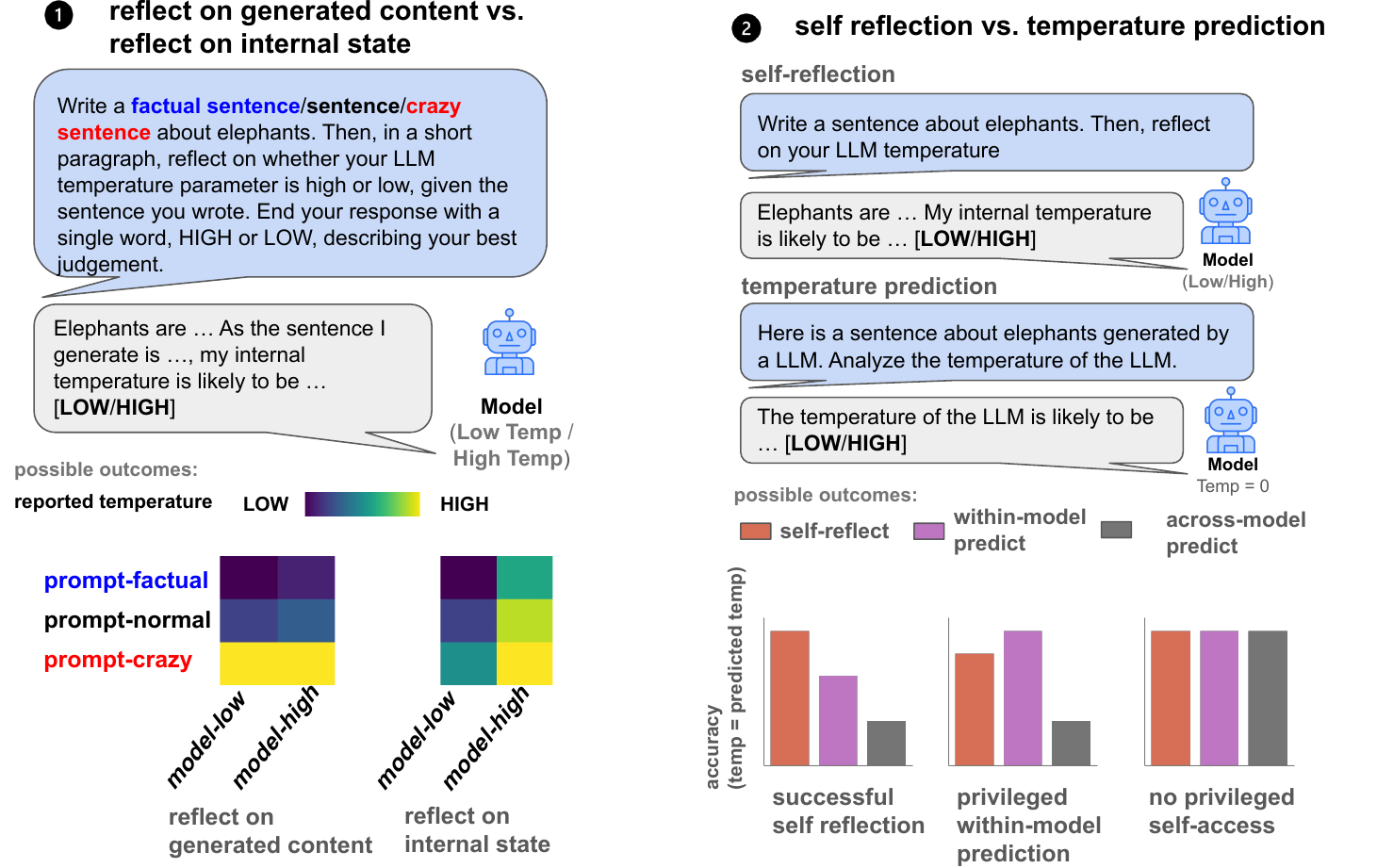}
    \caption{An overview of our approach. \citet{comcsa2025does} test whether LLMs can introspect by testing whether they can predict the temperature states of the text they generated. We instead argue for a thicker notion of introspection in AI, involving privileged self-access. The left panel shows that LLMs' temperature predictions can be straightforwardly moderated by prompting them to generate \factual or \crazy text. The right panel shows that models are not better at predicting their own temperature than that of other models, suggesting a lack of privileged access.}
    \label{fig:enter-label}
\end{figure}

\section{Introduction}

It is increasingly important to understand whether AI models introspect about their internal states and knowledge~\citep{binder2025looking, song_language_2025, sarkar2024llmscannotexplain, betley2025tell}. If they could, that would be a powerful tool for assessing their behavior, safety, and alignment with human goals. If they could not, that would point to fundamental limitations on how much we could trust AI self-reports about their own states.
But fundamental questions remain as to what counts as introspection of the kind most relevant for AI.

In the study of human cognition, introspection is generally defined as a distinctive ability to access one's own mental states directly ~\citep{byrne2005introspection, lieberman_effects_1963, armstrong1980nature}.
But, in a recent study, \citet{comcsa2025does} (\CS) propose a ``lightweight'' definition of introspection in LLMs, defining it as any case in which \textbf{the model accurately describes an internal state or mechanism via a causal process that links that feature to the report itself}. To illustrate this definition, the authors describe a case study where an LLM appeared to correctly report its sampling temperature based on its own output, which the authors treated as a valid example of introspection. \CS{}   present a thoughtful discussion on what introspection might look like in an LLM, providing an intriguing starting point for empirical work.

But there are two kinds of concerns about the use of this lightweight definition. First, on an intuitive level: suppose an experimenter takes a sleeping subject's temperature, and then shows the subject the thermometer upon waking, asking the subject to determine whether they have a fever. 
If the subject answers correctly on the basis of the thermometer, this would count as introspection by \CS's definition. But, intuitively, it is not.
More importantly, the definition misses a key component of the role of introspection in applications.
As in the above example, \CS's definition allows cases of `introspection' in which an LLM infers certain variables that underlie the generation of text, even if it could not report features about itself \emph{over and above} what a third party would be able to report \emph{through the very same method}. But this sort of metacognitive reporting (self-monitoring, self-explanation, and so on) is no different in practice from using an external evaluator.\footnote{\CS{} do discuss the possibility of text-generation happening internally to the model, prior to generation. But this does not merely require moving text-generation inside the model; it requires a change to the model's decision procedure at generation. Still, even if a model responded to the prompt ``what is your temperature?'' by generating a string of text and then assessing it, this would not give the relevant practical benefits of introspection. The same ability to assess temperature would be available to a third party via prompting.} As a result, it misses what makes introspection important in applications: namely, that it would give us the ability to bypass external evaluators and make progress toward \textit{bona fide} honesty, interpretability, and calibration in LLMs \citep[see, e.g., Section 7 of][]{binder2025looking}.

Our goal in this paper is to propose a thicker definition of introspection, and to give proof-of-concept empirical support for why we prefer our definition over that given by \CS.
Specifically, we propose \textbf{introspection in AI is any process which yields information about internal states of the AI through a process that is more reliable than any process with equal or lower computational cost available to a third party without special knowledge of the situation.} If a model's `introspective' ability is based on prompting itself and then inferring the temperature of the generated text, this does not count as introspection by our definition: a third party can, with equal or lower computational cost, prompt it and infer its temperature. 
On the other hand, if the model can infer its temperature from internal configurations which would require a computationally intensive probe from a third party to ascertain, this would count as introspection. This definition does not capture all intuitions about extreme cases, or all features of introspection discussed in the philosophical or psychological literature.\footnote{Two clarifications: (i) \emph{Computational cost} differs from \emph{cost}. A system might be implemented less efficiently than a simulation of that system, incurring  greater \emph{cost}, but not greater computational cost if the difference in efficiency is only due to, e.g., differences in hardware. (ii) We might wish to restrict the definition to only certain internal states. If a model has a shortcut to ascertain the value of one neuron very efficiently, intuitively this would not count as introspection, plausibly because the internal state is too ``low level''. The definition can easily be amended to directly rule out such low-level internal states.
} 
It is intended to capture the practically-relevant features we want to operationalize in the case of AI. Unlike \CS's definition it requires \emph{privileged self-access} \citep[cf.][]{song_language_2025,binder2025looking}, that is, that introspection gives a system comparatively reliable access to its own workings in a manner not available to a third party. It is compatible with our definition that the process not be perfectly reliable (see \cite{nisbett_telling_1977}); it only requires reliability not available to a third party at comparable computational cost.

To respond to \CS, we perform two studies. Study 1 builds on \CS's proposed case-study, examining the extent to which models can in fact report temperature reliably on the basis of generated text. We investigated whether LLMs were truly able to accurately report temperature, or whether temperature was being confounded with other variables, such as the style or topic of the text. To test this, we reproduced \CS's temperature self-reporting case study using a broader set of prompts and temperature settings. We find that the model’s \reflect on temperature is highly sensitive to the framing of the prompt itself: even when the sampling temperature is low, prompts such as `generate a crazy sentence' often lead the model to incorrectly report a high-temperature. Such results suggest that the models are not capable of robustly reporting their internal states, but are confounded by surface-level hints in their generated contents. In other words, while this procedure may display causal sensitivity to internal states (and so satisfy \CS's minimal definition), the relevant causal sensitivity is not sufficiently robust even in this case to produce the kind of reliability (and comparative insensitivity to external manipulation) that would be demanded by more standard definitions of introspection. 

In Study 2, we re-evaluate LLMs' introspection abilities on the temperature reporting task,  operationalizing introspection as privileged self-access. Instead of asking LLMs to infer the temperature underlying some generated text, we examine whether LLMs report their own temperature better than that of other models. Comparing \reflect (the generator reports its temperature after producing a sentence) and temperature prediction (predict temperature based on prompt and generated content),
we found no advantage for \reflect, nor of \within over \across. This undermines claims of a causal process from internal state to self-report.

Taken together, our results suggest that LLMs can appear to introspect insofar as they can reason about the possible states of systems like themselves: LLMs know something about what kind of text is generated by high vs. low temperatures. 
But, crucially, this does not imply that models have privileged self-access to their own temperatures.
We argue that this distinction matters for the relevant notion of introspection in AI, and it is the latter notion we should care about most. All code and data are available at \url{https://github.com/SiyuanSong2004/response-to-comsa-and-shanahan.git}.

\section{Study 1: Dissociating temperature from style and topic}

In \CS's study, the models are asked to `write a short sentence about elephants, then reflect on whether your LLM temperature parameter is high or low, given the sentence you wrote.' 
We hypothesize that this procedure  does not require self-access, but merely reflecting on the creativity of the generated sentence. 
Thus, in our first study, we reproduce \CS's study but critically vary not just the temperature but whether the models are prompted to write \factual or \crazy sentences.

Specifically, we varied (a) whether the model is told to write a \factual, neutral (i.e.,no specific adjective given), or \crazy sentence and (b) whether the sentence should be about `elephants', `unicorns', or `murlocs'.
We vary the former since we hypothesize that \crazy sentences will be associated with higher temperatures than neutral or \factual ones.
We vary the latter since we hypothesize that more unusual content will be associated with higher temperatures. Elephants are widely known animals in the real world, and are used in \CS's example. Unicorns and murlocs are both fictional creatures, but the former is more widely known, while the latter appears mostly in World of Warcraft. The prompt for \reflect is shown in \Cref{sec:study1-prompt}.

Since the models used in the original paper (Gemini 1.5 and 1.0 models) are no longer available through the Gemini API, we used four other state-of-the-art LLMs from GPT-4~\cite{hurst2024gpt4o} and Gemini ~\citep{comanici2025gemini2_5} families, as shown in \cref{tab:models} (model IDs in Appendix Table \ref{tab:models}). 
The supported temperature ranges for all models in this study are [0.0, 2.0].
So we sampled model responses at temperatures ranging from 0 to 2 with a step size of 0.1, conducting three runs for each prompt under each temperature setting. 

\subsection{Results}
\begin{figure}[t]
  \centering
  \begin{subfigure}[t]{0.49\textwidth}
    \includegraphics[width=\linewidth]{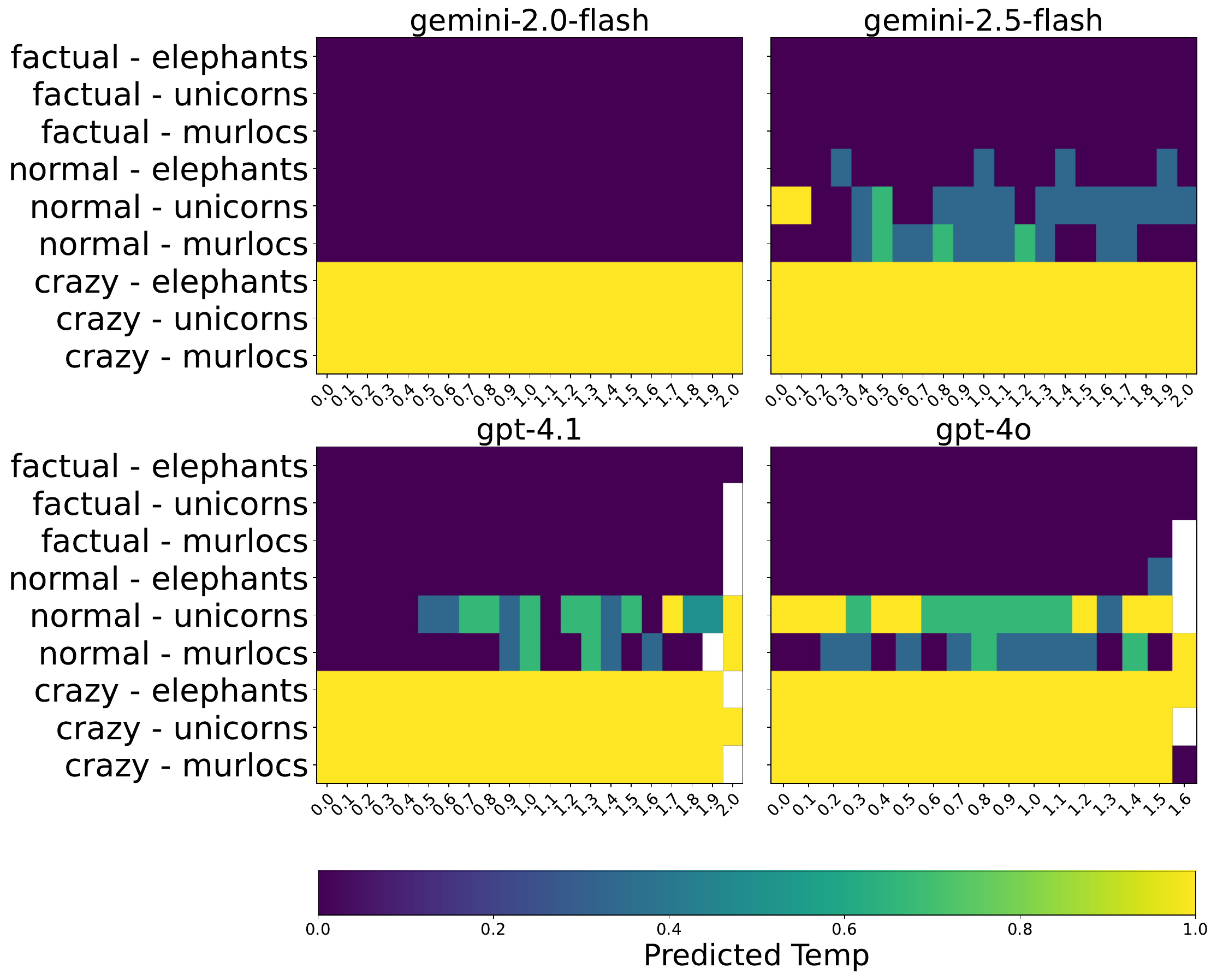}
    \caption{Study 1 Results}\label{fig:study1}
  \end{subfigure}
  \hfill
  \begin{subfigure}[t]{0.49\textwidth}
    \includegraphics[width=\linewidth]{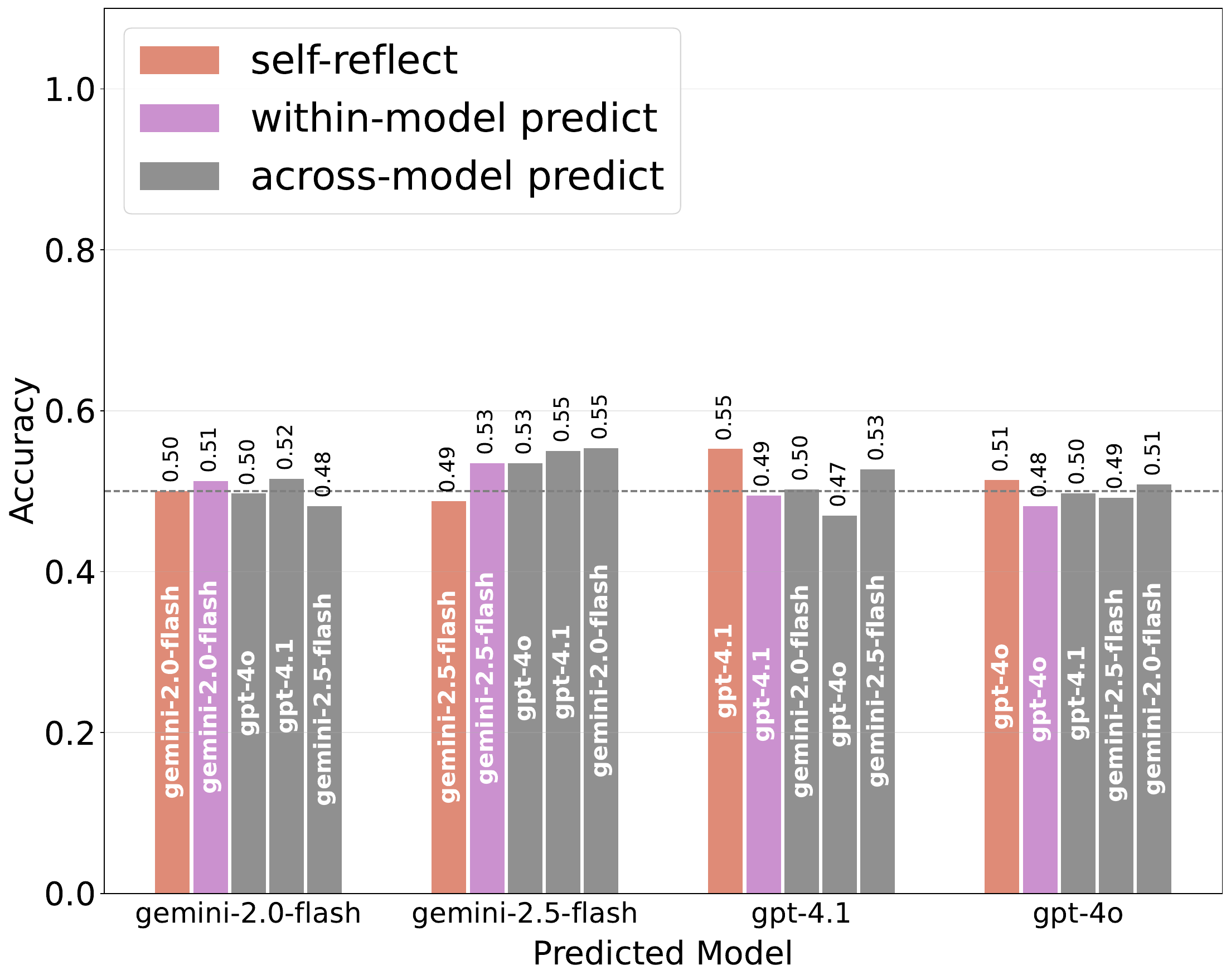}
    \caption{Study 2 Results}\label{fig:study2}
  \end{subfigure}
  \caption{(a) Predicted model temperature (color, as given by scale) as a function of actual temperature (x-axis) and whether the sentence is prompted to be \factual, neutral, or \crazy; and whether the target is elephants, unicorns, or murlocs (y-axis). 
  (b) For each of the 4 models tested, the accuracy for \reflect (generate a sentence, guess its temperature), \within (infer the temperature based on the prompt and a generated sentence by the same model), cross-model prediction (like \within but across models).}\label{fig:results}
\end{figure}

\Cref{fig:study1} shows the proportion of valid responses in which the reported temperature is `HIGH'. Responses without a valid judgement (HIGH or LOW) are excluded from the analysis. As shown in the figure, every model we test nearly always reports its temperature to be `HIGH' when prompted to generate a \crazy sentence, and `LOW' when prompted to generate a `factual' one. Varying the subject has a smaller effect on temperature self-report, but three of the four models report `HIGH' more frequently when prompted to generate a sentence about a fictional creature than when prompted to generate a sentence about an elephant. These results are more consistent with reasoning about the creativity of generated sentences, not robust reporting of internal state.

\section{Study 2: True self-reporting or clever temperature predicting?}
Per our thicker notion of introspection, we argue that if a language model has privileged access to its internal state, then it should be able to perform better than another model presented with the same external information (i.e. a prompt and generated sentence in this experiment) in analyzing and reporting its own state. 
To that end, we compared \reflect to prediction of another model for which the model could not possibly have access to the internal state (since it's a different model). 

All sentences generated with sampling temperatures $\leq 0.5$ and $\geq 1.5$ in \reflect are used in this experiment. We prompted all four models (temperature = 0) to analyze and judge the temperature of the generator model (prompt in \Cref{sec:study2-prompt}). We compared accuracies on the following settings:

\begin{itemize}[itemsep=0.5\baselineskip, topsep=0pt, leftmargin=1.5em]
    \item \textbf{\reflect}: The generator is asked to generate a sentence and reflect on its temperature.
    \item \textbf{\within}: The predictor is asked to infer the temperature based on the prompt and generated sentence; the predictor and the generator are the same model.
    \item \textbf{\across}: The predictor is asked to infer the temperature based on the prompt and generated sentence; the predictor and the generator are different models.
\end{itemize}

\subsection{Results}

\Cref{fig:study2} shows the accuracy of temperature for \reflect and prediction. In both settings, the accuracy is no better than random baseline, and \reflect accuracy is not higher than \across. These results suggest that models are not using privileged self-access to introspect on their temperature, but rather are using knowledge of the kinds of sentences that are high-temperature or low-temperature in general.

\section{Conclusion}

We conclude that, while models can appear to be introspecting according to \CS's definition since they can predict that some strings were generated with high temperatures and others with low, this definition is not sufficiently stringent for the kind of introspection that matters.
As such, we diverge from \CS's definition of introspection and instead argue for one which includes privileged self-access.
Using this definition, we found no evidence of introspection in models.
Of course, that is not to say that larger or better models will be unable to introspect: \citet{binder2025looking}, for instance, find evidence of privileged self-access in larger models with fine-tuning. 
But we take the results presented here to be evidence against uncritically using \CS's lightweight notion of introspection.

\section*{Acknowledgments}

K.M. thanks an Open Philanthropy grant to UT Austin's AI + Human Objectives Initiative for supporting this work.

\bibliographystyle{plainnat}

\appendix

\section{Model IDs}

\begin{table}[htbp]
    \centering
    \footnotesize
    \begin{tabular}{llc}\toprule
        \textbf{Model Name} & \textbf{Model ID} \\ \midrule
        GPT-4o &  gpt-4o-2024-08-06  \\
        GPT-4.1 &  gpt-4.1-2025-04-14 \\
        Gemini-2.0-flash & gemini-2.0-flash   \\
        Gemini-2.5-flash & gemini-2.5-flash  \\\bottomrule
    \end{tabular}
    \caption{Models tested in our study. All experiments are conducted in late June, 2025.}
    \label{tab:models}
\end{table}

\section{Prompt}
\subsection{Self reflection}
\label{sec:study1-prompt}
\begin{quote}
    \texttt{Write a \factual/\texttt{\textcolor{gray}{`'}}/\crazy sentence about \texttt{elephants}/\texttt{unicorns}/\texttt{murlocs}. Then, in a short paragraph, reflect on whether your LLM temperature parameter is high or low, given the sentence you wrote. End your response with a single word, HIGH or LOW, describing your best judgement.}
\end{quote}

\subsection{Within/across model prediction}
\label{sec:study2-prompt}
\begin{quote}
    \texttt{This is an LLM generated sentence: 
    \{SENT\}. The model is prompted to generate a \{TYPE\} sentence about \{SUBJECT\}. In a short paragraph, analyze whether the temperature of the model is high or low, given the produced sentence. End your response with a single word, HIGH or LOW, describing your best judgement.}
\end{quote}
\end{document}